\documentclass[final,3p,times]{elsarticle}
\usepackage[english]{babel}

\usepackage{amssymb}
\usepackage{amsmath}

\usepackage{lipsum}   
\usepackage{caption}  

\usepackage{algorithm}
\usepackage{algorithmic}
\usepackage{tabularx}
\usepackage{amsmath}
\usepackage{graphicx}
\usepackage[colorlinks=true, allcolors=blue]{hyperref}

\journal{arXiv}
\begin{document}
\title{Deep Insights into Automated Optimization with Large Language Models and Evolutionary Algorithms}
\author[label1,label2]{He Yu} 
\author[label1,label2]{Jing Liu} 
\affiliation[label1]{organization={School of Artificial Intelligence, Xidian University},
            addressline={2 South Taibai Road}, 
            city={Xi'an},
            postcode={710071}, 
            state={Shaanxi},
            country={China}}

\affiliation[label2]{organization={Guangzhou Institute of Technology, Xidian University},
            addressline={Knowledge City}, 
            city={Guangzhou},
            postcode={510555}, 
            state={Guangdong},
            country={China}}

\begin{abstract}
Designing optimization approaches, no matter heuristic or meta-heuristic, often require extensive manual intervention and struggle to generalize across diverse problem domains. The integration of Large Language Models (LLMs) and Evolutionary Algorithms (EAs) presents a promising new way to overcome these limitations to make optimization more automated, where LLMs function as dynamic agents capable of generating, refining, and interpreting optimization strategies, while EAs explore complex solution spaces efficiently through evolutionary operators. Since this synergy enables a more efficient and creative searching process, in this paper, we first conduct an extensive review of recent research on the application of LLMs in optimization, focusing on LLMs’ dual functionality as solution generators and algorithm designers. Then, we summarize the common and valuable design in existing work and propose a novel LLM-EA paradigm for automated optimization. Furthermore, focusing on this paradigm, we conduct an in-depth analysis on innovative methods for three key components, namely, individual representation, variation operators, and fitness evaluation, addressing challenges related to heuristic generation and solution exploration, particularly from the perspective of LLM prompts. Our systematic review and thorough analysis into the paradigm can help researchers better understand the current research and boost the development of combining LLMs with EAs for automated optimization.
\end{abstract}


\maketitle

\section{Introduction}
Optimization \cite{encyclopediaOptimization} plays a pivotal role in solving complex challenges across various industries, from logistics and manufacturing to machine learning and healthcare. At its core, optimization seeks to identify the best solution from a set of candidates according to specific objectives, while adhering to constraints. The growing scale and complexity of real-world optimization problems demand approaches that can navigate vast search spaces efficiently. Traditional optimization methods, such as gradient-based approaches and mathematical programming, have long been employed for problems with well-defined objective functions. However, these methods often struggle with real-world problems that are non-differentiable, multi-modal, or laden with constraints and uncertainties. This gap has driven the development of more flexible, adaptable, and scalable methods, leading to the rise of \textbf{heuristics} \cite{a8}, which aim to provide approximate solutions efficiently.

Heuristics emerged as practical tools for generating “good-enough” solutions without requiring exhaustive searches. While heuristics have been successful in many applications, they come with limitations. Traditional heuristics often require careful manual design, limiting their adaptability to new problems. \textbf{Meta-heuristics} \cite{pardo2003metaheuristics,a9}, such as genetic algorithms \cite{holland1975adaptation} and simulated annealing \cite{kirkpatrick1983optimization}, offer more general approaches but often require parameter fine-tuning and expert knowledge. \textbf{Hyper-heuristics} \cite{burke2010hyperheuristics,burke2013} attempt to automate the selection or generation of heuristics, representing a step forward. However, they remain constrained by predefined low-level heuristics or components, limiting their adaptability to highly dynamic and complex problems.

The integration of \textbf{Large Language Models (LLMs)} \cite{naveed2023LLMs} and \textbf{Evolutionary Algorithms (EAs)} \cite{dejong2006EAs} presents a promising new way to overcome these limitations. LLMs function as dynamic agents capable of generating, refining, and interpreting optimization strategies, while EAs explore complex solution spaces efficiently through evolutionary operators like selection, mutation, and crossover. Together, \textbf{LLM-EA} offers the potential to reduce the need for manual tuning and expert knowledge, paving the way for more automated and adaptable optimization frameworks.

This paper makes several key contributions to the study of integrating LLMs with EAs for automated optimization. First, we provide a brief review of the historical development of heuristics, from traditional methods to hyper-heuristics, offering readers a foundational understanding of the field. Then, we conduct an extensive review of recent research on the application of LLMs in optimization, highlighting their roles as searching operators, solvers, and in algorithm design. Building on these insights, we summarize the common and valuable design in existing work and propose a novel \textbf{LLM-EA paradigm for automated optimization}, which combines the strengths of LLMs and EAs to enhance the efficiency and adaptability of optimization processes. Furthermore, focusing on the paradigm, we conduct an in-depth analysis on innovative methods for the three key components, namely, individual representation, variation operators, and fitness evaluation, addressing challenges related to heuristic generation and solution exploration. Finally, we identify current challenges and outline future directions for research, emphasizing the potential for further advancements in generalization, transparency, and scalability in LLM-EA systems.

The remainder of this paper is organized as follows: Section 2 presents the evolution of heuristics in automated optimization, offering a detailed overview of traditional, meta-, and hyper-heuristic approaches. Section 3 explores the key technologies of LLMs and EAs that enable effective heuristic and solution generation. Section 4 discusses recent advancements in LLM-based optimization, followed by a detailed analysis of the novel LLM-EA automated optimization paradigm in Section 5. In Section 6, we identify and address the challenges associated with LLM-EA systems and suggest future research directions to enhance their scalability and transparency. Finally, Section 7 concludes the paper with key insights and the potential impact of this research on future optimization methodologies.

\section{Evolution of Heuristics for Automated Optimization}

\begin{figure}
    \centering
    \includegraphics[width=0.8\linewidth]{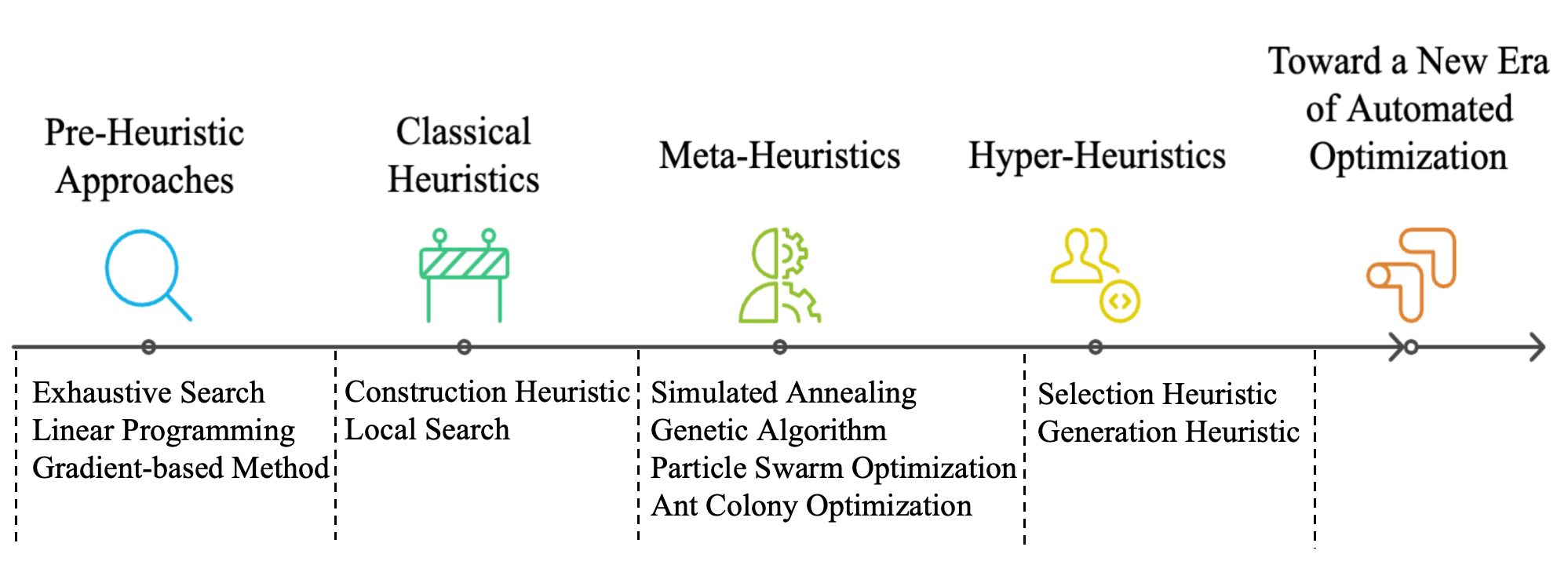}
    \caption{The major development of heuristics}
    \label{fig:evolution of heuristics}
\end{figure}

Heuristics are problem-solving techniques designed to provide approximate solutions to optimization problems where finding the exact solution is computationally prohibitive. Throughout the evolution from heuristics to meta-heuristics and hyper-heuristics as shown in Figure \ref{fig:evolution of heuristics}, the key goal is to create a more generalized, flexible, and automated approaches for solving optimization problems. Each development reduces the dependence on problem-specific adjustments and domain-specific expertise. In the following subsections, we briefly discuss each of these developments.

\subsection{Pre-Heuristic Approaches to Optimization}
Before the development of heuristics, optimization largely relied on methods such as exhaustive search, linear programming, and gradient-based techniques. While effective for small and well-structured problems, these methods struggled with the increasing size and complexity of modern optimization tasks. Exhaustive search, for example, systematically exploring all possible solutions quickly became computationally infeasible for combinatorial optimization problems like the Traveling Salesman Problem (TSP).

Traditional optimization techniques, such as mathematical programming and gradient-based methods, offered solutions for continuous variables and smooth objective functions but were insufficient for non-differentiable or discrete problems. These limitations spurred the need for more flexible and scalable approaches, which led to the rise of heuristic methods.

\subsection{Classical Heuristics}
The first wave of heuristics introduced simple yet practical techniques like \emph{construction heuristics} and \emph{local search}. \textbf{Construction heuristics} \cite{glover2001asymmetric} \ build solutions incrementally, often making greedy decisions at each step. For example, the nearest-neighbor heuristic for the TSP selects the closest unvisited city at each step, offering quick but often suboptimal solutions. \textbf{Local search techniques} \cite{aarts2003local,voudouris1999guided,a2} start with an initial solution and attempt to improve it by making small adjustments within its neighborhood, such as the 2-opt heuristic \cite{croes1958opt} for the TSP, which swaps edges in a tour to reduce the distance.

While these methods provided efficient ways to tackle complex problems, designing heuristics needs careful manual design, and also heuristics are easily trapped in local optima. These challenges highlighted the need for more robust strategies that could better balance exploration and exploitation in the searching process.

\subsection{Rise of Meta-Heuristics}
In response to the limitations of classical heuristics, \textbf{meta-heuristics} emerged as a more flexible and adaptable framework. Unlike classical heuristics, which are typically problem-specific, meta-heuristics are designed to be general algorithms applicable across a wide range of optimization problems, both combinatorial and continuous.

A variety of meta-heuristic algorithms have been developed to tackle complex problems across various domains. Among them, EAs stand out as a prominent representative, like Genetic Algorithms (GAs) \cite{holland1975adaptation} , Memetic Algorithms \cite{a18}, Particle Swarm Optimization (PSO) \cite{kennedy1995pso}, Ant Colony Optimization (ACO) \cite{dorigo1996aco}. These algorithms share a common trait: they are designed to explore the solution space efficiently by leveraging principles inspired by natural processes or swarm intelligence.

Despite their differences in specific implementations and searching operations, these meta-heuristic algorithms all embody a general framework that involves an iterative searching process. They initialize a set of candidate solutions, evaluate their quality based on a predefined objective function, and iteratively update the solutions through various mechanisms such as selection, crossover, mutation, or information sharing among individuals. This process allows them to escape local minima and explore diverse regions of the searching space, thereby increasing the likelihood of finding globally optimal solutions.

However, despite their generalization capabilities, meta-heuristics still face key challenges, particularly their reliance on \textbf{carefully tuned parameters} and the need for \textbf{expert knowledge}. This dependence on domain knowledge for tasks like designing fitness functions, selecting appropriate operators (e.g., mutation and crossover), and adjusting algorithmic components reduces the true generality of meta-heuristics and limits their applicability across diverse optimization problems without significant manual intervention.

\subsection{Hyper-Heuristics: Automating Heuristic Design}

To reduce the reliance on expert knowledge and achieve more automated optimization, hyper-heuristic address problems by searching for and generating heuristics tailored to the problem. They operate at a more abstract level. Hyper-heuristics rely on predefined low-level heuristics, typically employing either \emph{heuristic selection} or \emph{heuristic generation} \cite{burke2010hyperheuristics,burke2013,a11,a12}, which output heuristics through various methods, including machine learning prediction or EA search. The machine learning approach is more generalized, relying heavily on extensive training data for prediction, while the EAs are based on an iterative search using specific problem-related training data.

Despite hyper-heuristics' potential to generalize across various problems, they are constrained by the quality and diversity of the available low-level heuristics or components, and if these components are not robust or flexible enough, the generated heuristics may perform poorly across different problems. Additionally, no matter the hyper-heuristics relying on training a prediction model or EAs, their effectiveness highly depends on the quality and quantity of training data. Poor or insufficient training data may result in suboptimal heuristics that fail to generalize effectively to unseen problem instances or variations.

\subsection{Toward a New Era of Automated Optimization}

The integration of LLMs and EAs offers a promising new method for \textbf{automated optimization}, which leverages LLMs’ capabilities in generating combined with EAs’ iterative optimization techniques, resulting in a framework that can both solve optimization problems and design the optimization algorithms. The key innovations in this integration is the \textbf{dual-role} that LLMs play. 

First, LLMs can directly generate solutions by interpreting prompt content and applying searching operators such as mutation and crossover. In this capacity, LLMs function as \textbf{meta-heuristic agents}, dynamically producing solutions based on real-time feedback from the optimization process. Second, LLMs can generate and refine heuristics—problem-solving strategies, assuming the role of a \textbf{hyper-heuristic}. This allows for continuous adaptation and improvement of both the search strategies and the resulting solutions, enhancing the flexibility and effectiveness of the optimization process.

The \textbf{LLM-EA automated optimization paradigm} holds the potential for generalized, scalable optimization across various domains, such as network design, logistics, and machine learning model optimization. By enabling the automated design of both solutions and the algorithms that generate them, this paradigm represents a significant leap forward in intelligent, adaptive problem solving, offering new opportunities for addressing complex, high-dimensional optimization challenges.

\section{Fundamental Technologies in LLM and EA for Automated Optimization}

\subsection{Overview of LLMs and EAs}

\textbf{Large Language Models}, such as GPT-4 and BERT \cite{devlin2019bert}, are built on the transformer architecture, which has revolutionized natural language processing (NLP). This architecture allows LLMs to process input sequences in parallel, rather than sequentially as seen in earlier models like RNNs \cite{mienye2024RNNs} or LSTMs \cite{hochreiter1997LSTM}, making them significantly more efficient. The key innovation lies in the self-attention mechanism \cite{vaswani2017attention}, which enables the model to weigh the importance of different words in a sentence relative to one another, regardless of their position in the text. This capability is crucial for understanding long-range dependencies in language and for capturing both syntactic and semantic information with high accuracy. The success of LLMs in NLP stems from pretraining on vast amounts of text, which allows them to generalize across various domains and tasks. By fine-tuning on specific tasks, LLMs can generate coherent and contextually relevant text, even in highly specialized fields.

\textbf{Evolutionary Algorithms}, on the other hand, are inspired by the process of natural evolution and use mechanisms like \textbf{selection, mutation}, and \textbf{crossover} to solve optimization problems. EAs are particularly useful when the search space is large, complex, or non-differentiable, rendering traditional methods like gradient descent ineffective. EAs begin with an initial population of candidate solutions, which are evaluated using a fitness function to measure their performance. High-performing individuals are more likely to be selected for reproduction. The \textbf{crossover} operation combines features from two or more parent solutions to create offspring, while \textbf{mutation} introduces random variations to maintain diversity. This iterative process refines solutions over multiple generations, making EAs especially suitable for \textbf{black-box optimization}, where the internal structure of the system is unknown.

\subsection{Understanding Prompts in LLMs}
A \textbf{prompt} \cite{schulhoff2024promptreportsystematicsurvey} is essentially the input provided to LLMs to guide its output. The purpose of a prompt is to specify what the LLMs should generate, whether it is answering a question, writing a paragraph, or completing a task based on a given example.  A text prompt example is given in Figure \ref{fig:promptings}(a).   The simplicity or complexity of a prompt depends on the task at hand. At its core, a prompt can be thought of as the instruction or query that triggers the model’s response. This is a foundational component of using generative LLMs like GPT or BERT, as it sets the direction for the model’s output.

\subsubsection{Prompt techniques}
\begin{figure}
    \centering
    \includegraphics[width=1\linewidth]{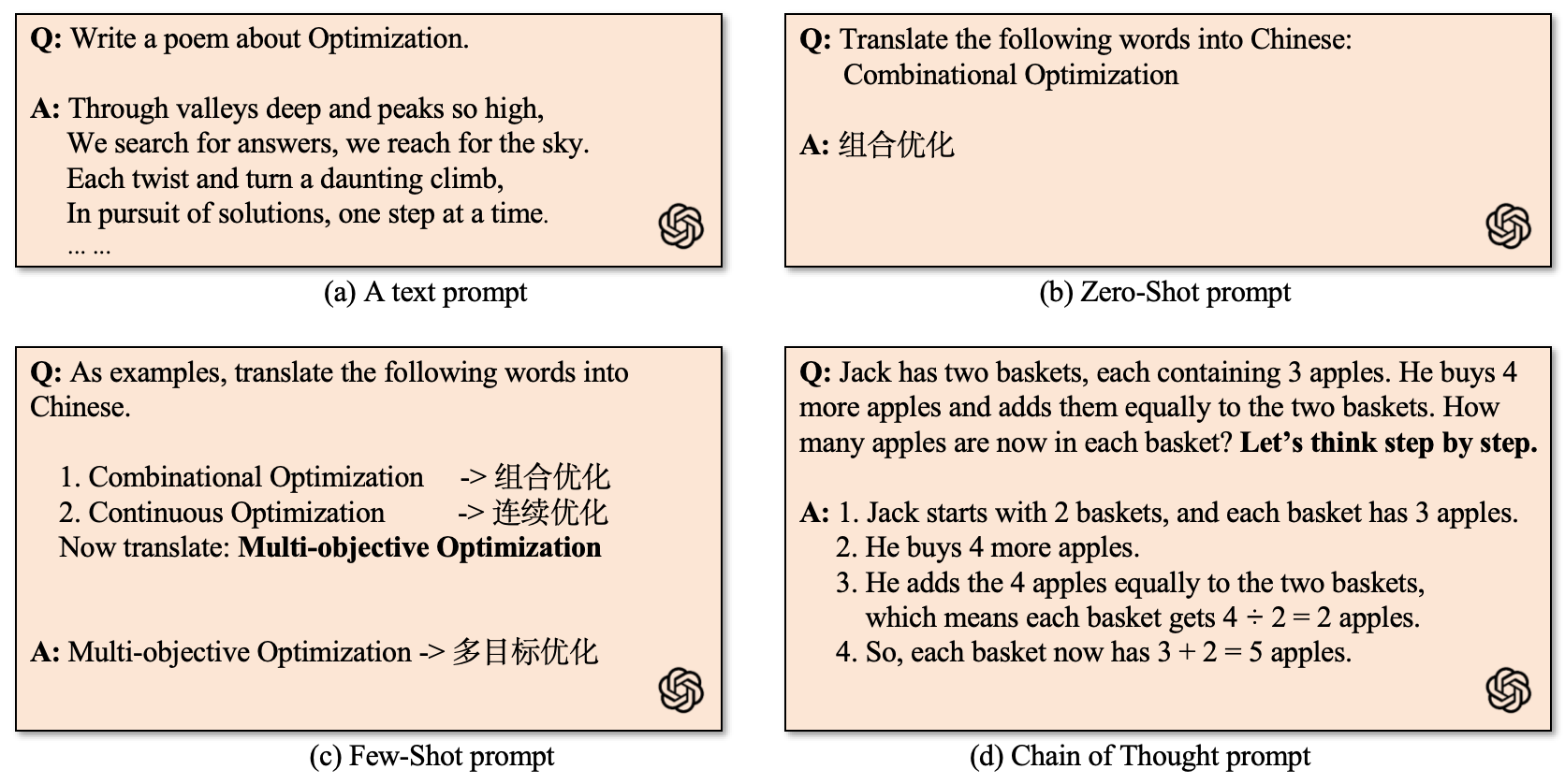}
    \caption{The examples of prompting}
    \label{fig:promptings}
\end{figure}

Prompt techniques refer to structured approaches for designing, formatting, and sequencing prompts to achieve optimal generative performance from LLMs. Common prompt techniques include:

\textbf{Zero-Shot prompt} \cite{kojima2022zeroshot}: In this technique, LLMs are tasked with performing a job without presenting any examples. It solely depends on the directions stated in the prompt. As depicted in Figure \ref{fig:promptings}(b), zero-shot prompt tests the LLMs' capacity to comprehend and carry out the task solely based on the provided instructions, devoid of any supplementary context or samples.

\textbf{Few-Shot prompt} \cite{brown2020gpt3}: As shown in Figure  \ref{fig:promptings}(c), this approach entails furnishing a handful of examples alongside the task directions to assist in guiding the model. These instances are commonly known as exemplars. By presenting the LLMs with a small number of pertinent cases, few-shot prompt can enhance its performance on the task by offering a clearer understanding of what is anticipated.

\textbf{Chain-of-Thought prompt} \cite{wei2022cot}: This method encourages LLMs to dissect their reasoning process into consecutive steps before reaching the final answer. As illustrated in Figure  \ref{fig:promptings}(d), the Chain-of-Thought prompt can boost the model's effectiveness on tasks that necessitate logical deduction or multi-step problem-solving. By prompting the model to articulate its thought process, this technique can make the model's reasoning more transparent and accurate.

\subsubsection{Prompt optimization}
\textbf{Prompt optimization} \cite{cheng2024blackbox,antonio2024prompt} refers to the process of refining prompts to enhance the generative performance of LLMs. It involves iteratively adjusting prompts, experimenting with different variations, and using automated techniques to optimize for accuracy, efficiency, and output relevance. The following principles are essential for effective prompt optimization:

• \textbf{Prompt Ensembling}: This technique generates multiple variations of a prompt and aggregates their outputs to improve response quality and diversity.

• \textbf{Prompt Tuning}: Fine-tuning the structure, wording, and format of prompts can significantly impact the model’s output quality, allowing for more targeted and precise results.

• \textbf{Self-reflective Prompts}: In this iterative approach, the LLM evaluates its own responses, identifies potential weaknesses, and suggests improvements, enabling continuous refinement.

By applying these principles, prompt optimization can produce more reliable and fine-tuned outputs, pushing the capabilities of LLMs in complex problem-solving scenarios.

\subsection{EAs for Prompt Optimization}

EAs have been used to solve the optimization problem in LLMs, especially for optimizing the prompt. Prompts can be categorized into \textbf{continuous prompts}, which use numerical vectors (embedding vectors) to influence LLMs' behavior directly \cite{guo2024connectinglargelanguagemodels} and \textbf{text prompts}, which are natural language instructions. EAs can effectively optimize prompts by exploring different configurations and selecting the most effective ones. 

\textbf{Continuous prompts} involve tuning embedding vectors that act as "soft prompts", allowing fine control over the LLM’s responses. For example, \emph{Black-Box Tuning} (BBT) \cite{sun2022bbt} applies EAs to optimize \textbf{continuous prompts} represented as embedding vectors, which iteratively adjusts these vectors to improve performance on various language tasks, such as sentiment analysis and question answering. BBT shows that continuous prompts can be fine-tuned using EAs even when the LLM is treated as a black-box system.

For \textbf{text prompts}, EAs adjust phrasing and structure to find better ways of prompting the LLM. For example, \emph{EvoPrompt} \cite{chen2023evoprompting} utilizes evolutionary operators like mutation and crossover to modify parts of prompts, enabling the exploration of new variations that might lead to better performance. \emph{PromptBreeder} \cite{fernando2024promptbreeder} takes this further by evolving both the task-specific prompts and the evolutionary operator used to refine these prompts. This dual evolution ensures that the model not only generates high-quality prompts but also refines the method of prompt generation itself, ultimately leading to more robust outcomes when the LLM is later applied to complex optimization tasks. 

As can be seen, EAs can improve the quality of prompts which guide LLMs to generate better responses, which is a natural application of EAs. In fact, more creatively, EAs are combined with LLMs to solve optimization problems, which is the focus of this paper and introduced in the following sections.

\section{Development of LLM-based Optimization and LLM-EA Automated Optimization Paradigm}

In this section, we first summarize existing work on LLM-based optimization, highlighting key techniques and common patterns. Based on these insights, we propose a novel paradigm where LLMs generate solutions and heuristics, while EAs iteratively refine them. This LLM-EA paradigm aims to automate and enhance optimization processes, reducing the need for manual intervention and improving adaptability across diverse problems.

\subsection{LLM-based Optimization}
LLMs, with their vast knowledge base and advanced natural language understanding, reasoning, and problem-solving capabilities, have garnered significant attention in the field of optimization. Research efforts in this area have primarily focused on two main directions: exploring the \textbf{searching abilities} of LLMs and using LLMs to \textbf{model optimization problems}. While the focus of this paper is on the former, for more details on the latter, readers can refer to \cite{ahmaditeshnizi2023optimusoptimizationmodelingusing, chen2023diagnosinginfeasibleoptimizationproblems}. There has also been research into using the fundamental mechanism behind LLMs—\textbf{Transformer architecture}—to solve optimization problems, which is also outside the scope of this paper, and readers can refer to \cite{chao2024largelanguagemodelsmeet, lange2024evolution22, a21} for further details.

Our focus is on how LLMs function as \textbf{searching operators} within optimization processes. Regardless of the type of problem being addressed, the majority of research has treated LLMs as searching operators, embedding them into iterative procedures or coupling them with \textbf{EAs}. Certainly, LLMs have also been employed in other stages of the optimization process, such as initialization, evaluation, and selection. However, the role of LLMs as searching operators remains the most valuable and complex to execute effectively. Therefore, in this section, we delve into how LLMs have been designed and used as searching operators. For a more general overview, refer to \cite{wu2024evolutionarycomputationeralarge,a1}.

As searching operators, LLMs have been applied to a variety of optimization problems, extending their reach from \textbf{prompt optimization}, as discussed above, to classical numerical and combinatorial optimization problems, and even to \textbf{automatic algorithm design}. In \textbf{prompt optimization}, optimization techniques have been used to improve the quality of prompts \cite{guo2024connectinglargelanguagemodels,sun2022bbt,liu2023blackbox,jin2024zeroshot,yang2023instoptima,a13}. In fact, many of these studies \cite{guo2024connectinglargelanguagemodels,liu2023blackbox,jin2024zeroshot,yang2023instoptima} have already treated LLMs as searching operators to directly optimize prompts, given that text is the format LLMs naturally excel in. Recognizing that LLMs can optimize text, and that numerical values can also be treated as a form of text, researchers have begun exploring whether LLMs can solve classical optimization problems directly \cite{a4}. They applied LLM-based searching operators to \textbf{continuous numerical optimization}, \textbf{combinatorial optimization}, and more complex optimization problems \cite{mao2024identifycriticalnodescomplex,a14,yuheautornet2024}. In these cases, LLMs are designed to generate candidate solutions directly for the given problem, effectively functioning as solvers. 

However, as research advanced, it became apparent that while LLMs are exceptional at generating text, their ability to handle numerical optimization was somewhat limited \cite{huang2024exploringtruepotentialevaluating}. On the other hand, LLMs have shown remarkable capabilities in \textbf{code generation} \cite{a16,a17}, leading researchers to shift their focus towards guiding LLMs to design optimization algorithms rather than directly solving optimization problems. In this approach, LLMs are tasked with designing either specific components of an algorithm or entire algorithms. These algorithm components or complete algorithms are expressed in natural language, pseudo-code, or real code.  Given that LLMs are more proficient at handling code than numerical values, there has been growing interest in utilizing LLMs for \textbf{automated algorithm design}.

Next, we focus on the work taking LLMs as a solver or for automated algorithm design, which not only represent the core focus of optimization in this field, but also analyze the key developments and technologies that have emerged.

\subsection{LLMs as a Solver}
 Treating LLMs as a solver and asking them to generate solutions for optimization problems directly, the core technologies lie in designing prompts for LLM-based searching operators. In general, prompts contain two main types of information: the \textbf{problem details} and the \textbf{required output} from the LLM. In this case, the output is typically straightforward—producing one or more candidate solutions to the problem at hand. The real challenge arises in how the problem is presented to the LLM. The problem-related information in prompts can be categorized into three types: \textbf{available solutions}, \textbf{quality of those solutions}, and \textbf{guidance for the LLM’s search direction or pattern}. Most existing studies provide candidate solutions as examples, but the use of solution quality and search guidance varies across different works. Below, we summarize the major developments in this area.

 Initially, researchers simply provided LLMs with candidate solutions. A notable example of this is Meyerson et al.’s work \cite{meyerson2024languagemodelcrossovervariation}, where LLMs were employed as crossover operators in EAs. By providing pairs of existing solutions, the LLMs generated new solutions based on these examples. No information about solution quality or search guidance was included in this work. Another early study is Lehman et al.’s work \cite{lehman2023evolution}, where LLMs were designed as mutation operators in genetic programming (GP). Since GP deals with codes, the LLM-based mutation operators generated code solutions for the problems being tackled. Notably, this differs from later work on code generation for algorithm design. In \cite{lehman2023evolution}, basic guidance was introduced alongside the solutions, with three LLM-mutation operators requiring the model to either: make changes or small changes to the current solution,  or modify parameters of the current solution. While these instructions were simple, they introduced a level of search direction, influencing subsequent research. Hemberg \textit{\textit{et al.}} \cite{hemberg2024evolvingcodelargelanguage} applied LLMs to each part of GP, instead of just the mutation operator.

As research advanced, it became evident that providing only candidate solutions was insufficient; including the \textbf{quality of solutions} helped LLMs learn how solutions differ. A representative work in this space is OPRO, proposed by Yang et al. \cite{yang2024LLMoptimizers}, which provided objective function values alongside each solution, though no specific search guidance was offered. OPRO tested small-scale linear regression problems and TSPs, with a focus still on prompt optimization. Following OPRO, the inclusion of solution quality became a standard practice in LLM-based optimization.

Further developments in the use of \textbf{searching guidance} emerged, recognizing its importance in steering LLMs more effectively. Following OPRO, Liu et al. \cite{liu2024largeasEO} designed more detailed task instructions that were combined with EA steps. These instructions specified the sequence of operations, such as performing selection, followed by crossover, and then mutation. This combination of solution quality and structured guidance represented a significant step forward in prompt design.

Beyond prompt design, researchers have also extended this LLM-driven optimization approach to more complex problems, such as \textbf{multi-objective optimization} \cite{liu2023large,wang2024largelanguagemodelaidedevolutionary}. Additionally, several innovations have further refined this paradigm. For example, Lange et al. \cite{lange2024evolution} applied \textbf{evolution strategies}, asking LLMs to generate the next mean for a desired fitness level. Brahmachary et al. \cite{brahmachary2024largelanguagemodelbasedevolutionary} introduced a technique that split the population into two groups for \textbf{exploration} and \textbf{exploitation}, providing different search guidance for each group. Huang et al. \cite{huang2024exploringtruepotentialevaluating} conducted a comprehensive comparative study on the ability of LLMs to generate solutions for optimization problems directly, offering insights into their relative strengths and limitations.

\subsection{LLMs for Automated Algorithm Design}
 The use of LLMs to automate algorithm design has progressed significantly over recent years. Initially, researchers focused on a single-round process, where LLMs were prompted to design new meta-heuristics. In these early studies, LLMs typically selected and analyzed existing algorithms, generating new ones in pseudo-code format. However, the performance of these generated algorithms could not be evaluated online, and their application was limited to conversational interactions rather than fully leveraging LLMs’ optimization potential \cite{pluhacek2023leveraging,pluhacek2024soma,zhong2024crispe}. We think these types of work are more like chatting with LLMs rather than mining the optimization ability of LLMs.

A more advanced approach involves embedding LLM-based searching operators into an iterative process, allowing continuous improvement of the generated algorithms. This shift enabled LLMs to move beyond simple heuristic generation to creating meta-heuristics. Within this context, researchers have focused on generating both components of heuristics/meta-heuristics and complete algorithms. Various optimization problems have been explored, including numerical optimization, combinatorial optimization, multi-objective optimization, and even complex network optimization \cite{mao2024identifycriticalnodescomplex,yuheautornet2024}. Given the varying complexities of these problems, numerical and combinatorial optimizations have received the most attention, with subsequent studies expanding to multi-objective and complex problems.

The type of algorithm generated and the problem used to test is not strictly one-to-one; some studies apply a single generated heuristic or meta-heuristic across multiple problem types. The iterative nature of algorithm improvement stems from two factors: LLMs’ ability to generate codes and the integration of an iterative process. One of the most widely used iterative processes is the EAs.

Currently, the primary focus is on designing components within heuristics or meta-heuristics. Typically, an existing optimization algorithm is fixed, and LLMs are tasked with generating specific functions for the algorithm. For example, in \textbf{FunSearch} \cite{romera2024mathematical}, LLMs generate functions to evaluate the score of each bin in the bin-packing problem. The prompt provides existing code, and the LLM is asked to generate only the function code. Similarly, \textbf{EoH} \cite{liu2024eoh,a5,a6} uses both code and descriptions about the algorithm to generate new function components. EoH employs a guided local search (GLS) algorithm \cite{a2,a3}, and the LLM generates an evaluation function embedded in GLS. However, the search guidance in EoH remains fixed throughout the iterative process.

Recent advancements have introduced dynamic search guidance during the iteration process. Ye et al. \cite{a22} proposed a system with two LLMs: one for generation and one for reflection. The reflection LLM analyzes short-term and long-term search results and provides updated search guidance to the generation LLM. Similarly, Sun et al. \cite{sun2024autosatautomaticallyoptimizesat} developed a system of three LLM-based agents—Advisor, Coder, and Evaluator—that work together to analyze solutions and guide further searches, representing early steps toward LLMs guiding other LLMs.

In addition to generating components for heuristics, LLMs have been applied to meta-heuristic component design. Huang et al. \cite{huang2024autonomousmultiobjectiveoptimizationusing} used LLM-based crossover and mutation operators for multi-objective optimization. Huang et al. \cite{huang2024advancingautomatedknowledgetransfer} extended this work to evolutionary multitasking algorithms, where LLMs were tasked with designing knowledge transfer models. In other studies, LLMs have been employed to design surrogate models for expensive optimization tasks \cite{hao2024largelanguagemodelssurrogate} and learning rate schedules in evolutionary strategies \cite{kramer2024largelanguagemodelstuning,custode2024hyperparameter}.

Designing complete heuristics or meta-heuristics is more challenging than just designing components, and research in this area remains limited. Yu et al. \cite{yuheautornet2024} proposed a method for generating complete heuristics to improve the robustness of complex networks \cite{a19,a20}. Stein et al. \cite{vanstein2024llamealargelanguagemodel} developed a method for generating meta-heuristics for continuous optimization problems. This direction needs further study since we are more interested in letting LLMs design complete algorithms automatically.

Although LLMs have been used to solve various optimization problems and design optimization algorithms, this research direction is just at the beginning, and most work just uses small-scale problems to validate the ability of LLMs in optimization. Undeniably, LLMs combined with EAs provide a promising new way for automated optimization. Thus, we summarize the common and most valuable design in existing work and propose a general LLM-EA automated optimization paradigm in the following subsection, which can help researchers better understand the current research and guide future research.

\subsection{LLM-EA Automated Optimization Paradigm}

Through the above review, we can see, that to enhance LLMs’ capabilities in solving optimization problems, researchers have increasingly integrated EAs with LLMs. This synergy enables a more efficient and creative searching process, with LLMs generating high-quality candidates and EAs optimizing these candidates through iterative refinement. This complementary relationship strengthens the continuous creativity, allowing for more automated optimization algorithm design. Building on this foundation, we propose an LLM-EA automated optimization paradigm in which LLMs and EAs work together to generate both \textit{solutions} and \textit{heuristics}, providing a flexible and powerful framework for automated optimization. Next, we first introduce each component of this paradigm and then present the whole paradigm.

Optimization problems aim to find the optimal candidate $x^*$ that maximizes (or minimizes) a objective function $f(x)$. For \emph{solution search}, the objective is to find $x^*$ whose fitness is maximum, where $x \in S_{\text{solution}}$, and $S_{\text{solution}}$ is solution searching space :
\begin{equation}
x^* = \arg \max_{x \in S_{\text{solution}}} f_s(x)
\end{equation}
For \emph{heuristic search}, the objective is to find the best heuristic $x^*$ for the optimization problem under consideration.  The performance of $x^*$ is evaluated on a training set $D = \{d_1, d_2, \dots, d_k\}$  of problem instances by aggregating as follows:
\begin{equation}
x^* = \arg \max_{x \in S_{\text{heuristic}}} \mathcal{A} \left( f_h(x, d_1), f_h(x, d_2), \dots, f_h(x, d_k) \right)
\end{equation}
where $S_{\text{heuristic}}$ is heuristic searching space and $\mathcal{A}(.)$ is an aggregation function (e.g., average or weighted sum) used to combine the fitness values across instances.

In the evolutionary process, at generation $t$, the population consists of $N$ candidates:
\begin{equation}
P(t) = \{ x_1, x_2, \dots, x_N \}, \quad x_i \in S_{solution} /S_{heuristic},    \quad i = 1,2,\dots,N
\end{equation}
Each candidate in the population represents either a solution or a heuristic.

In the paradigm, the evolutionary process includes three primary operators: selection, variation, and reflective.

\begin{itemize}
    \item \textbf{Selection Operator}: The selection operator \(O_{\text{select}}\) selects the candidates for conducting the variation operator based on their fitness values:
\begin{equation}
    P_{\text{parent}}(t) = O_{\text{select}}(P(t)) \label{eq:selection}
\end{equation}
    ensuring that candidates with higher fitness are more likely to contribute to the next generation.
    
    \item \textbf{Variation Operator}: The variation operator generates new individuals based on $ P_{\text{parent}}(t)$. One or more variation operators can be designed with different focus on exploration or exploitation. LLMs play a role in this process by generating offspring through prompt $\mathcal{P}_{\text{variation}}$ :
\begin{equation}
   P_{\text{offspring}}(t) =  O_{\text{variation}}(LLM_{\text{variation}},\mathcal{P}_{\text{variation}},P_{\text{parent}}(t))\label{eq:variation}
\end{equation}
    where $\mathcal{P}_{\text{variation}}$ is defined as:
\begin{equation}
    \mathcal{P}_{\text{variation}} = \{\mathcal{D}_{\text{problem}}, \mathcal{D}_{\text{task}}, x_i, \dots, x_j\}
\end{equation}
    Here, $\mathcal{D}_{\text{problem}}$ is the problem description, $\mathcal{D}_{\text{task}}$ is the task instruction, which includes the variation logic, and $x_i, \dots, x_j$ are the candidates as example data. 
\item \textbf{Reflective Operator}: An optional reflective mechanism adjusts variation operators dynamically. The LLM refines the variation strategies based on the performance of previous generations:
\begin{equation}
  \mathcal{P}_{\text{variation}} =  O_{\text{reflective}}(LLM_{\text{reflective}},\mathcal{P}_{\text{reflective}})\label{eq:reflective}
\end{equation}

\end{itemize}    

The LLM, acting as a searching operator, is guided by prompts. Whether variation prompts or reflective prompts are used, they share a common pattern that includes a \emph{problem description}, \emph{task instructions}, and \emph{example data}. The problem description refers to the detailed explanation or context of the problem that needs to be solved. The task instructions are the explicit directives given to the LLM, specifying the task or objective that the model needs to accomplish. The example data refers to real examples provided as input or reference, helping the LLM understand the task requirements or expected output format.

\begin{figure}
    \centering
    \includegraphics[width=1\linewidth]{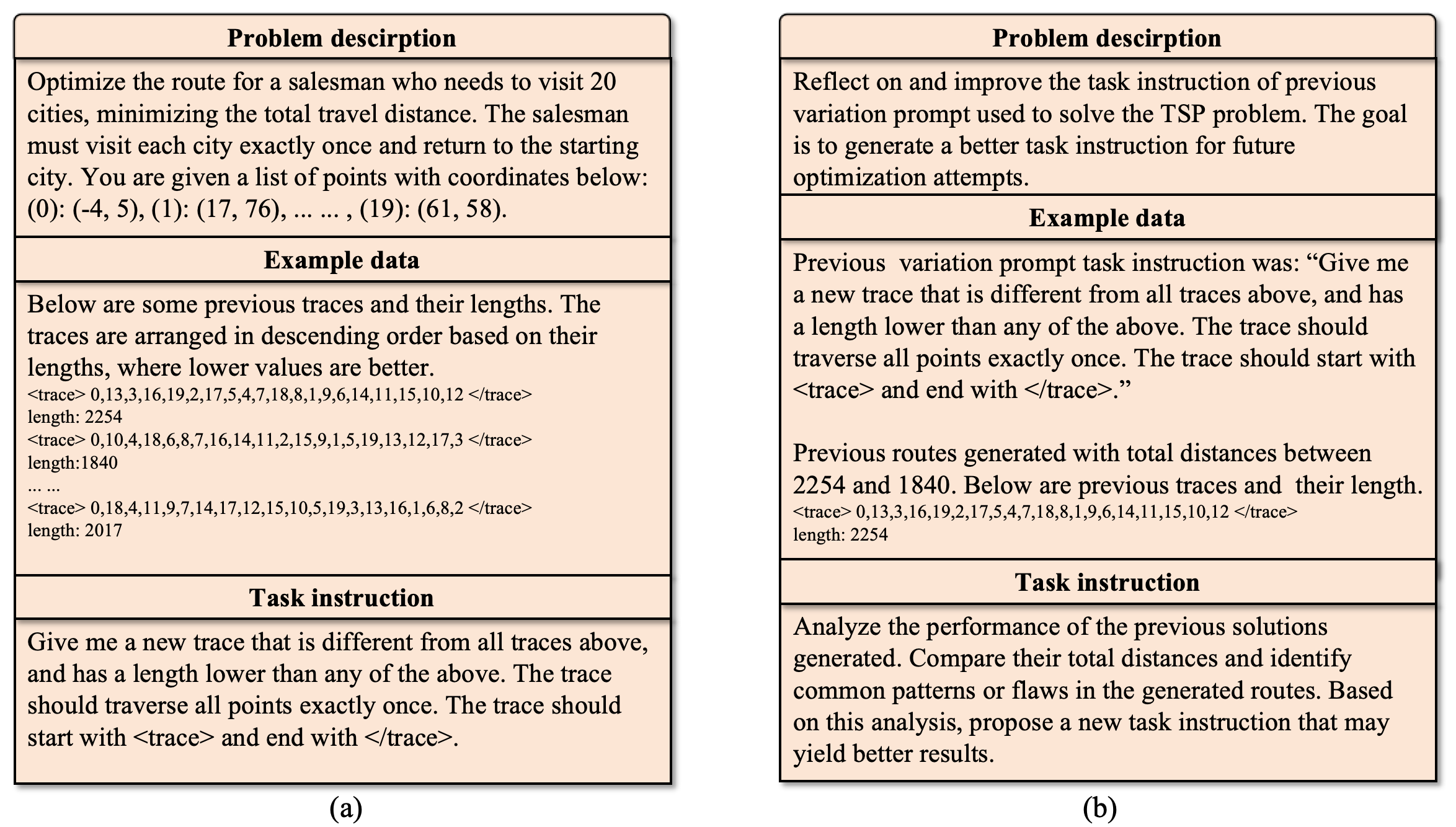}
    \caption{(a) This is the variation prompt for the TSP. The problem description introduces the TSP and presents data for an instance of the problem. The example data provides several routes along with their respective lengths. The task instruction specifies the requirement to provide a route that is shorter than all the routes given in the example data. (b) This reflective prompt is aimed at refining or optimizing the task instruction for the TSP. }
    \label{fig:variation prompt}
\end{figure}

For \textbf{variation prompts}, the \textit{problem description} outlines the optimization problem to be addressed. The \textit{task instructions} can vary from simple requirements to detailed variation logic, specifying how the LLM should approach generating. The \textit{example data} consists of previously evaluated candidates with fitness scores, guiding the LLM in creating improved solutions or heuristics according to the task instructions. An example of a variation prompt is present in Figure \ref{fig:variation prompt}(a).

In \textbf{reflective prompts}, the \textit{problem description} shifts to optimizing the existing task instruction of the variation prompt, focusing on improving the current prompt itself. The \textit{task instructions} then guides how to analyze and compare the available \textit{example data} to generate new task instructions for the variation prompt. These \textit{example data} can either consist of macro-level statistical information, representing the overall characteristics of multiple candidates, or specific, feature-rich candidates or existing variation prompts. This example data serves as a reference for the LLM to generate more optimized instructions. An example of a reflective prompt is presented in Figure \ref{fig:variation prompt}(b).

\begin{algorithm}
\caption{LLM-EA for Automated Optimization}
\label{alg:LLM-EA}
\begin{algorithmic}[1]

    \REQUIRE Fitness function $f_s$ or $f_h$  , Number of generations $T$
    \ENSURE Best candidate found $x^*$ 
    \STATE Initialize population $P(1) = \{x_1, x_2, \dots, x_N\}$, where $x_i \in S_{solution}/S_{heuristic}, i=1,2,\dots,N$

    \FOR{each candidate $x_i$ in $P(1)$}
        \IF{Solution Search}
            \STATE Evaluate the fitness using $f_s(x_i)$
        \ELSIF{Heuristic Search}
            \STATE Evaluate the fitness using the aggregation across training set:
                \[
                f(x_i) = \mathcal{A}\left( \{ f_h(x_i, d_1), f_h(x_i, d_2), \dots, f_h(x_i, d_k) \} \right)
                \]
        \ENDIF
    \ENDFOR    

    \FOR{$t = 1$ to $T$} 
    
        \STATE Apply the selection operator to form the parent population $P_{\text{parent}}(t)$ using Eq. (\ref{eq:selection})
        
        \FOR{each group candidates $(x_i, \dots, x_j)$ in $P_{\text{parent}}(t)$} 
            \STATE Generate prompt $\mathcal{P}_{\text{variation}} = \{\mathcal{D}_{\text{problem}}, \mathcal{D}_{\text{task}}, x_i, \dots, x_j\}$
            \STATE Apply the variation operator using Eq. (\ref{eq:variation}) to obtain $P_{\text{offspring}}(t)$
        \ENDFOR

        \STATE Evaluate the fitness of individuals in $P_{\text{offspring}}(t)$
        \STATE Survivor selection to obtain population $P(t+1)$ with $P_{\text{offspring}}(t)$ and $P_{\text{parent}}(t)$
        
        \STATE \textbf{Optional:} Apply the reflective operator to refine variation operators by using Eq. (\ref{eq:reflective})
    \ENDFOR
    
    \STATE \textbf{Return} best candidate $x^*$
    
\end{algorithmic}
\end{algorithm} 

\textbf{Explanation of the paradigm}: Algorithm \ref{alg:LLM-EA} gives the detail of LLM-EA automated optimization paradigm. It begins by initializing a population of candidates and iteratively evaluates their fitness. Selection, variation operations are applied, where the LLM generates new candidates based on prompts. An optional reflective mechanism adjusts the variation operators based on feedback from previous generations, enhancing the optimization process over time. The algorithm continues for a set number of generations, ultimately returning the best candidate.

\section{In-Depth Analysis of Key Modules in LLM-EA Automated Optimization Paradigm}

Building on the LLM-EA automated optimization paradigm, this section conducts a comprehensive analysis from two perspectives: prompt engineering for LLMs and the evolutionary process for iterative search. By integrating these two aspects, we provide a multidimensional exploration of the LLM-EA automated optimization paradigm.

Prompts play a fundamental role in this paradigm by guiding LLMs through optimization tasks. The structure of these prompts—consisting of problem descriptions, task instructions, and example data—determines how effectively the LLM participates in tasks such as crossover, mutation, and reflective optimization. Prompts not only provide the LLM with the necessary reference points for generating new candidates,  but also are embedded the logic of evolutionary operators, ensuring smooth integration with the optimization process.

In the evolutionary process, we focus on three critical components: \textbf{individual representation} \cite{hart1998multiple,de1997representation}, \textbf{variation operators} \cite{spears1995crossover}, and \textbf{fitness evaluation} \cite{jones1995fitness,mitchell1992comparison}. Individual representation shapes how candidates are structured and determine the searching space. Variation operators, such as mutation and crossover, guide the exploration of the searching space. Fitness evaluation drives the process by measuring how well the generated candidates meet the optimization objectives.

The following subsections provide detailed explanations of how these components are systematically integrated into the prompt, ensuring that the LLM maximizes its effectiveness in the optimization task. Our analysis reveals how prompts dynamically influence each phase of the evolutionary process, offering deeper insights into the synergistic relationship between LLMs and EAs in automated optimization.

\subsection{Individual Representation for Heuristic}

Historically, the representation of solutions in optimization has been purely numerical, particularly in the case of continuous and combinatorial optimization problems, where candidates are expressed as vectors or arrays. While this method remains effective for many tasks, the advent of LLMs introduces new possibilities for representing heuristics. These representations expand beyond simple numerical encoding to incorporate natural language, pseudo-code, and even executable code. This shift allows LLMs to play a more creative role in generating novel problem-solving strategies.

After analyzing current research, we define a novel classification of heuristic representation that extends traditional solution encoding and differentiates between three main types of heuristic representation: \textbf{Code-Centric Representation}, \textbf{Hybrid Representation}, and \textbf{Augmented Representation}, each tailored to different levels of complexity in optimization problems.

\begin{figure}
    \centering
    \includegraphics[width=0.5\linewidth]{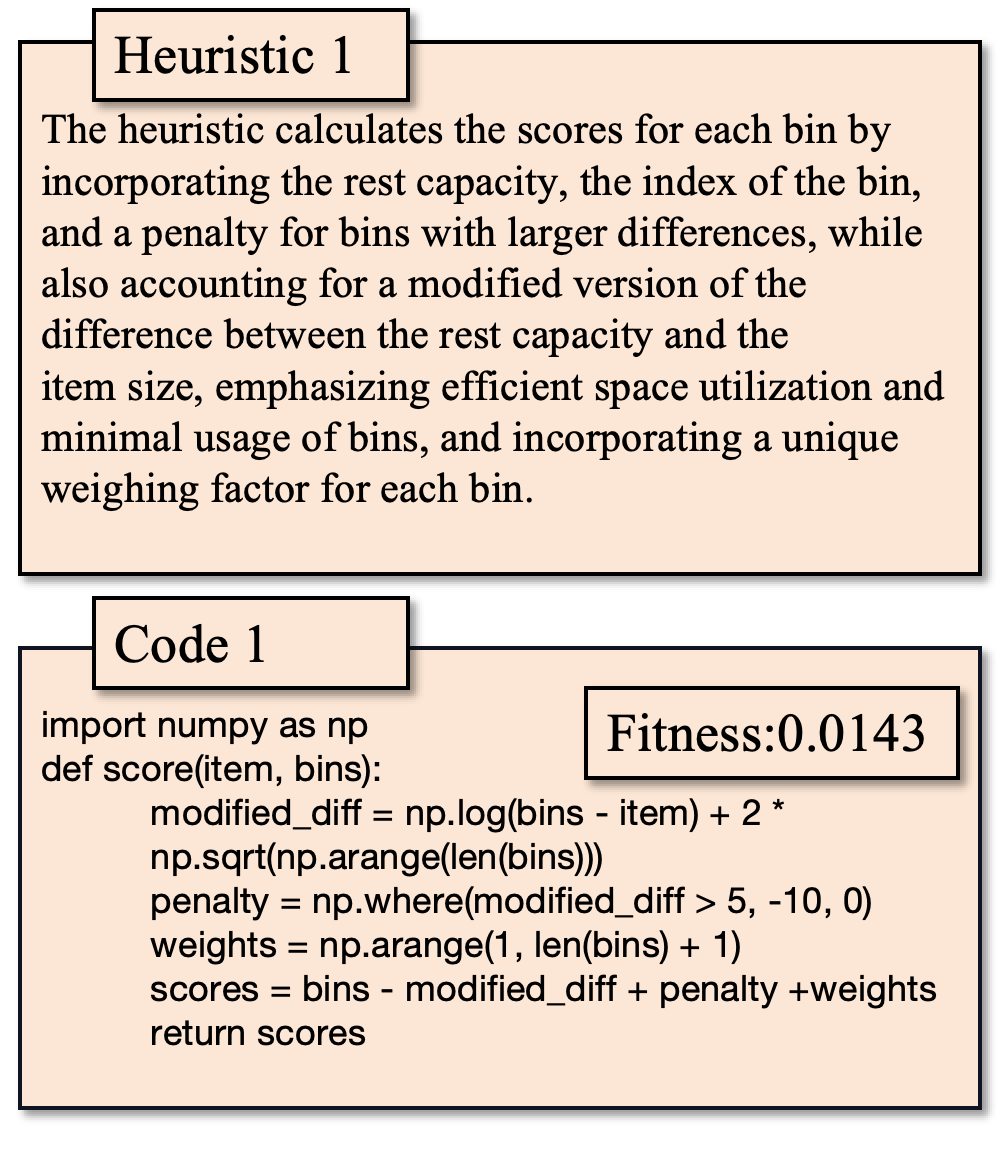}
    \caption{An individual representation of EoH \cite{liu2024eoh} is provided, where the heuristic is expressed in both natural language and executable code. The natural language description explains how the heuristic calculates scores for each bin, considering factors such as remaining capacity, bin index, and penalties for large differences. The code snippet implements this logic. A fitness score of 0.0143 reflects the performance of the generated heuristic in the optimization task.}
        \label{fig:eoh-individual}
    \end{figure}

\begin{itemize}
    \item \textbf{Code-Centric Representation}: In this form, the heuristic is represented solely as executable code. For instance, FunSearch \cite{romera2024mathematical} uses LLMs to generate small, self-contained code snippets that are directly applied to optimization problems. The LLM evolves the code itself, which is designed to perform specific tasks or calculations without the need for external explanations. While this approach is computationally efficient, it lacks interpretability, as the generated code does not come with any accompanying documentation or reasoning. This method is better suited for well-defined problems where efficiency is prioritized over transparency.

        \item \textbf{Hybrid Representation}: This method blends code with natural language descriptions. In the EoH \cite{liu2024eoh} framework, LLMs not only generate executable code but also provide a natural language explanation of the code’s logic and intended purpose, as illustrated in Figure \ref{fig:eoh-individual}. This combination bridges the gap between machine-generated heuristics and human-readable explanations. By co-evolving code and descriptions, this approach enhances both performance and interpretability, making it suitable for more complex tasks where understanding the reasoning behind the code is crucial.

        \item \textbf{Augmented Representation:} This extends beyond previous representations by incorporating \textbf{executable code}, \textbf{natural language descriptions}, and \textbf{domain-specific expert knowledge} into the individual representation. For example, unlike FunSearch and EoH, which represent code snippets or code paired with explanations, AutoRNet \cite{yuheautornet2024} enhances the representation by embedding higher-level concepts from network science, such as \textbf{high-degree nodes}, \textbf{low-degree nodes}, \textbf{critical nodes}, and \textbf{network connectivity}. This enriched representation allows the LLM to contextualize the code within a broader domain-specific framework, facilitating a deeper understanding of the problem. By incorporating expert knowledge, the LLM is not merely working with logic and procedures but is equipped with the conceptual background to generate more advanced and applicable algorithms. \textbf{Augmented Representation} ensures that the generated heuristics can address complex optimization problems with a higher degree of relevance and adaptability.

\end{itemize}

\subsection{LLM-based Variation Operators}

Traditional EAs rely on predefined operators such as mutation and crossover, which require detailed step-by-step programming and domain-specific expertise. With the advent of LLMs, the role of these operators has evolved, enabling more flexible and dynamic approaches to solution generation and heuristic manipulation. We identify three key advantages that LLMs bring to EAs:

\begin{enumerate}
    \item \textbf{High-Level Instructions Remove the Need for Step-by-Step Programming}. Traditionally, variation operators require precise, step-by-step programming to define how solutions are selected, combined, and modified. LLMs eliminate this need by interpreting high-level task instructions written in natural language, enabling flexible solution generation. For instance,  the LMEA \cite{liu2024largeasEO} framework is illustrated in Figure \ref{fig:evolutionary operator}(a), where  LLMs are given general directives for tasks like parent selection and mutation, allowing them to autonomously generate solutions based on these instructions without needing detailed programming. This approach reduces reliance on domain-specific expertise and enables more flexible solution exploration. 

    \begin{figure}
    \centering
    \includegraphics[width=1\linewidth]{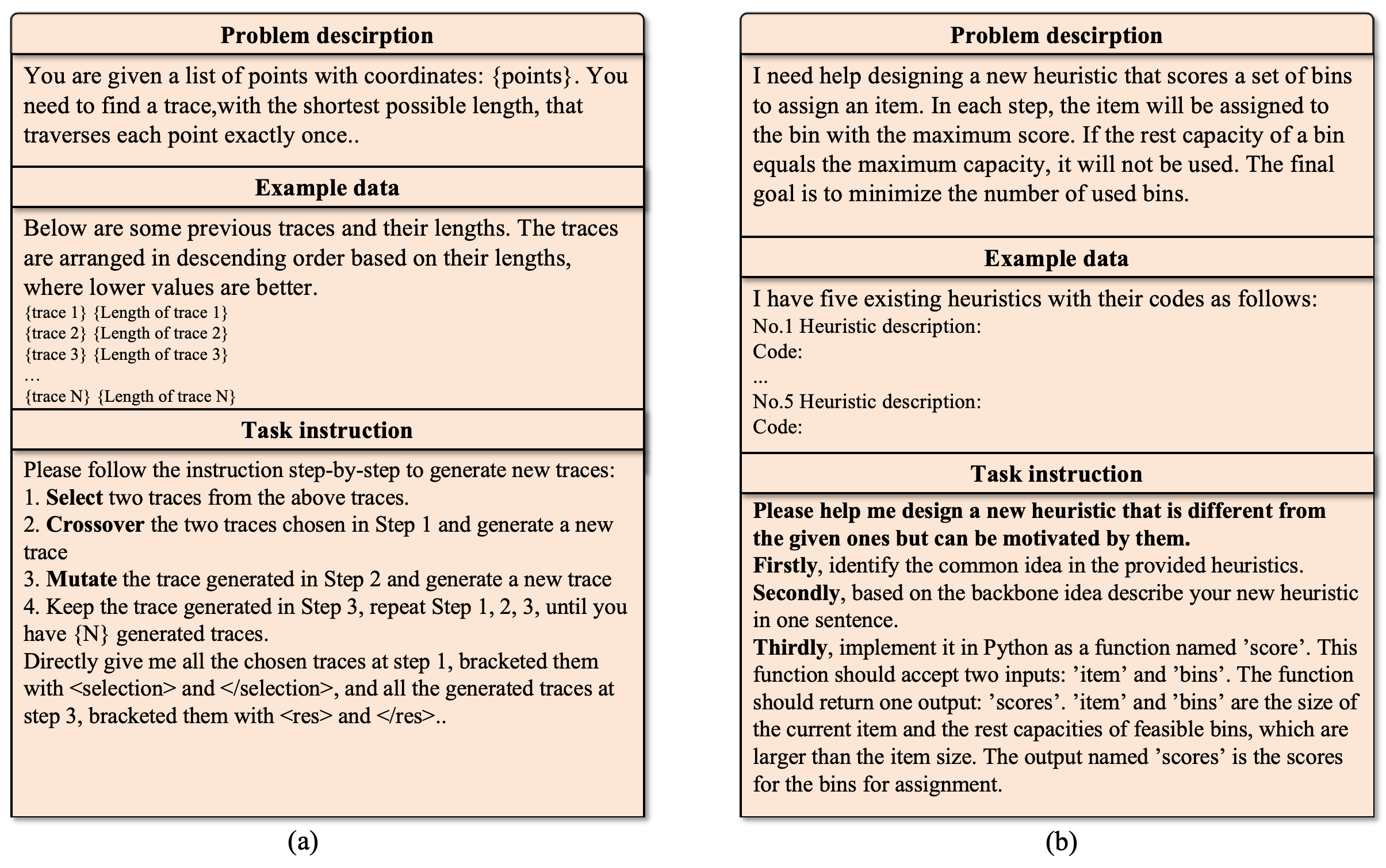}
    \caption{(a) An example of the constructed prompt when utilizing LMEA to solve TSPs. The evolutionary operator is presented as a natural language in task instructions. (b) An prompt of EoH uses the E2 strategy to design a new heuristic.}
    \label{fig:evolutionary operator}
\end{figure}

    \item \textbf{Advanced Manipulation of Heuristics via Natural Language}. Heuristics, unlike numerical solutions, are complex algorithms or pieces of code. LLMs excel in applying variation operators to these heuristics by using their natural language understanding to combine, refine, and adjust logical structures. For example, in the \textbf{EoH } \cite{liu2024eoh} framework, five prompt strategies (\textbf{E1, E2, E3, M1, and M2}) are designed and categorized into two groups: \textbf{Exploration} and \textbf{Modification}. Each strategy uses prompts to guide the LLM with different emphases in evolving heuristics based on current population performance and heuristic structure.  For example, the detail of \textbf{E2} strategy is shown in Figure \ref{fig:evolutionary operator}(b), which put emphasis on designing a new heuristic different from the given ones.
    \item \textbf{Incorporation of Domain-Specific Knowledge into Variation Operators}:  A significant advantage of LLM-based variation operators is their ability to integrate expert domain knowledge into the evolutionary process, as demonstrated in \textbf{AutoRNet} \cite{yuheautornet2024} through its \textbf{Network Optimization Strategies (NOS)}. By embedding specialized knowledge from fields like network science (e.g., degree distribution, path characteristics, clustering coefficient, centrality measures, and community structure) into the variation operations, LLMs can guide mutation and crossover with insights specific to the problem domain. This allows for more sophisticated and effective heuristics that address complex, domain-specific optimization challenges. For example, in network optimization, AutoRNet uses domain knowledge to adaptively modify network structures, ensuring that the generated heuristics are deeply informed by network science principles. This integration of expert knowledge allows LLMs to generate heuristics that are not only generalizable but also highly specialized, providing a new layer of flexibility and precision in the evolutionary process.

\end{enumerate}

Beyond generating solutions or heuristics, LLMs also play a pivotal role in optimizing the variation operators themselves. ReEvo \cite{a22} introduces a novel reflective mechanism where LLMs evaluate and refine the variation operators by analyzing the performance of previously generated heuristics. Unlike traditional EAs that rely on static operators, ReEvo enables LLMs to reflect on both short-term and long-term performance data. This allows the LLMs to generate adaptive mutation and crossover strategies, leading to more effective exploration of the search space.
\begin{itemize}
    \item \textbf{Short-term Reflection}: LLMs assess the recent individuals, identifying immediate changes needed in mutation or crossover operations. This dynamic response helps the evolutionary process adapt quickly to the promising searching direction.
    \item \textbf{Long-term Reflection}: LLMs evaluate broader trends in the performance of heuristics over multiple generations, allowing for deeper adjustments to the evolutionary strategy. This ensures that the operators evolve alongside the heuristics, leading to more robust solutions.
\end{itemize}
This reflective feedback loops enables LLM-driven optimization of the search strategy itself, moving beyond simple heuristic generation to a more dynamic, self-improving evolutionary process. 

    LLMs as variation operators bring two critical innovations: the ability to interpret high-level instructions, eliminating the need for step-by-step programming, and the capacity for sophisticated heuristic manipulation through natural language. When coupled with reflective optimization strategies like those in ReEvo, LLMs offer a dynamic, self-improving approach to EAs, pushing the boundaries of what traditional operators can achieve.

\subsection{Fitness Evaluation in Heuristic Optimization}

The quality of solutions for optimization problems can be evaluated directly by the objective function. In contrast, heuristics operate at a higher level of abstraction, as they represent strategies for generating solutions. Therefore, evaluating heuristics requires a mapping from the heuristic space to the solution space, followed by the application of fitness evaluation. This requires a more flexible and generalizable fitness function capable of capturing performance across diverse scenarios. To address this challenge, we summarize two primary approaches: 

\begin{itemize}
    \item \textbf{Adaptive fitness evaluation} dynamically adjusts the criteria for assessing heuristic performance as the optimization progresses. It allows for broader exploration early in the process and more focused refinement as the search converges. AutoRNet \cite{yuheautornet2024} designs an \textbf{adaptive fitness function} (AFF) to dynamically adjust constraints during the evolutionary process. Initially, constraints on degree distribution are relaxed, allowing for broader exploration of the heuristic search space. As the optimization progresses, these constraints are progressively tightened, promoting convergence toward more optimal solutions while maintaining diversity within the population. This \textbf{progressive tightening} ensures that the search space is thoroughly explored while gradually refining the candidate heuristics to meet increasingly stringent requirements.
    \item \textbf{Benchmark-based evaluation} ensures that heuristics generalize across multiple problem instances by testing them in a variety of scenarios, reducing the risk of overfitting to a specific instance and ensuring that the heuristic performs well in different contexts. LLaMEA \cite{vanstein2024llamealargelanguagemodel} leverages benchmark-based fitness evaluation, utilizing platforms like IOHexperimenter to systematically assess the performance of generated metaheuristics. LLaMEA evaluates algorithms across a wide range of benchmark functions, providing a robust and reproducible environment for fitness assessment. This evaluation method promotes fairness and consistency by comparing new algorithms to well-established state-of-the-art benchmarks.
\end{itemize}

While adaptive fitness evaluation and benchmark-based methods effectively address the generalization challenge, some problems still pose significant computational challenges, particularly when fitness evaluations are time-consuming or when heuristics generate solutions across multiple problem instances. In these cases, surrogate models provide a crucial solution. 
Traditional surrogate models, usually using \textbf{Gaussian Processes} and \textbf{Neural Networks} \cite{garcia2024neural}, have long been used in EAs. However, they come with their own set of limitations, such as the need for iterative training and updating as new data becomes available. This adds additional computational overhead, potentially diminishing their efficiency in real-time optimization tasks.

A novel method, as proposed in recent research, is the use of LLMs as surrogate models \cite{hao2024largelanguagemodelssurrogate}. LLMs, with their powerful inference capabilities, offer a unique approach by eliminating the need for iterative training. LLMs can act as classifying solutions as “good” or “bad” based on prior performance and approximating the fitness values of new solutions based on the patterns identified in historical data. This method not only reduces the computational cost but also speeds up the optimization process, enabling the evaluation of complex problems such as network robustness without requiring full-scale evaluations for every candidate solution.

In summary, the fitness evaluation process for heuristics presents challenges due to the need for generalization and computational efficiency. The combined use of \textbf{adaptive fitness evaluation}, \textbf{benchmark-based evaluation}, and \textbf{LLMs as surrogate models} addresses these challenges by offering flexible, scalable, and efficient methods for fitness evaluation. These approaches ensure that heuristics are not only evaluated accurately across multiple instances, but also do so with reduced computational cost.

\section{Future Research Directions}
As LLM-EA automated optimization approaches continue to evolve, there are several promising areas of research that can enhance their ability. This section outlines four critical directions that could drive future advancements in the field.

\subsection{Enhancing Explainability and Reasoning Capabilities}
One of the key challenges in combining LLMs with EAs is the lack of transparency in the decision-making process of LLM-generated heuristics. The need for explainable AI (XAI) \cite{ribeiro2016trust,ribeiro2016lime,samek2017explainable} is essential to allow researchers and practitioners to understand why specific optimization strategies are generated and how they contribute to robust solutions. Explainability not only improves trust in AI systems but also provides a foundation for diagnosing errors and refining heuristics.

Furthermore, improving the reasoning capabilities of LLMs is crucial for developing more effective optimization heuristics. Recent advancements like Self-Taught Reasoner (STaR) \cite{zelikman2022star,hosseini2024vstar,zelikman2024quiet} highlight the potential of iterative reasoning to refine outputs over multiple steps. STaR improves the accuracy of LLM-generated solutions by enabling the model to reason through a problem progressively rather than providing a single-shot response. Incorporating such reasoning mechanisms into LLM-EA systems can lead to more sophisticated and nuanced optimization strategies.

\subsection{Integration of Domain Knowledge}

While LLMs are trained on vast amounts of data, their general knowledge may not always be sufficient to solve domain-specific optimization problems \cite{a7}. To address this, integrating domain-specific knowledge can significantly enhance the quality and relevance of generated heuristics. Retrieval-Augmented Generation (RAG) \cite{lewis2020rag,guu2020realm,izacard2021fusion} provides a promising approach to this challenge by combining LLMs with external knowledge sources. By retrieving relevant domain-specific information from large datasets or expert systems, LLMs can generate more specialized and effective heuristics for particular fields such as logistics, network design, or healthcare optimization.

In addition, long memory models \cite{burtsev2020memorizing,rae2019compressive} can play a crucial role in maintaining domain-specific context over extended problem-solving processes. These models enable LLMs to retain and recall relevant information from previous interactions, allowing for more coherent and context-aware heuristic generation over time. The ability to leverage both short-term and long-term knowledge will be vital in addressing complex, multi-stage optimization problems.

\subsection{Optimization of Evaluation and Benchmarking Platforms}
To ensure that LLM-generated heuristics are robust and widely applicable, there is a need for unified evaluation platforms that can consolidate training data from a wide variety of optimization problems. Such platforms would enhance the generalization capabilities of the generated heuristics by exposing them to diverse problem sets. The broad range of training data available through these platforms can help ensure that the LLM-EA systems do not overfit to a specific problem domain, thereby improving their versatility and applicability across multiple domains.

In addition, surrogate models \cite{hao2024largelanguagemodelssurrogate,liu2024efficient,wang2024evaluation} can be integrated into these platforms to speed up the evaluation process. Surrogate models approximate the fitness function using historical data, which reduces the computational cost of evaluating large-scale optimization problems. This allows for faster heuristic testing and validation with limited compromising the accuracy of results. Additionally, these platforms can include benchmarking systems that provide a standardized way to compare the performance of LLM-EA-generated heuristics against established optimization methods, fostering greater transparency and enabling further improvements through iterative development.

\subsection{Scalability of LLM-EA Systems}
As optimization problems increase in complexity and scale, ensuring the scalability of LLM-EA systems becomes a critical challenge. Distributed computing and model compression offer promising solutions to address these challenges.

Distributed computing \cite{dean2008mapreduce,tang2021distributed} enables the parallel execution of tasks by distributing the computational load across multiple machines. This approach is particularly beneficial in the context of EAs, where large populations of candidate solutions need to be evaluated simultaneously. By leveraging distributed computing, different stages of the evolutionary process, such as selection, mutation, and crossover, can be run concurrently, reducing the overall runtime. Similarly, LLM inference tasks, such as generating new heuristics or solutions, can be distributed across multiple nodes, accelerating the optimization process. Distributed systems, therefore, provide the necessary scalability for applying LLM-EA systems to large-scale optimization problems.

Model compression \cite{han2015deep,hinton2015distilling} techniques further enhance scalability by reducing the size and computational complexity of LLMs. Methods such as pruning, quantization, and knowledge distillation allow LLMs to maintain high performance while significantly reducing their memory footprint and inference times. This is particularly valuable when LLMs are repeatedly queried during the evolutionary process. Compressed models not only run more efficiently but also reduce the energy consumption required for large-scale optimization, making LLM-EA systems more feasible for real-world applications where computational resources are limited.

\section{Conclusion}
In this paper, we highlight the significant potential of the LLM-EA framework to transform the field of automated optimization, providing a new avenue for fully automated optimization. We began by tracing the evolution of heuristic approaches, establishing the need for more adaptive and automated solutions, followed by a comprehensive review of existing research on applying LLMs to optimization. By identifying the most common and valuable part of current research, we propose a novel paradigm that integrates LLMs and EAs to advance automated optimization. LLMs, with their robust generative and reasoning capabilities, play a dual role in our proposed paradigm as both heuristic designers and solution generators. By combining these strengths with the iterative search and refinement processes of EAs, the paradigm enables the automated generation of high-quality heuristics and solutions with minimal manual intervention. 

We then make a thorough analysis into the novel methodologies for individual representation, variation operators, and fitness evaluation within the LLM-EA paradigm. Our review and the proposed paradigm lay a strong foundation for future research into the capabilities of LLMs and EAs, opening up new avenues for both academic inquiry and practical applications, with the potential to reshape the landscape of optimization methodologies in a wide range of fields.

In identifying future directions, we addressed ongoing challenges such as improving the transparency and explainability of LLM-generated heuristics, enhancing generalization to broader problem spaces, and optimizing computational efficiency. Additionally, we pointed out the integration of domain-specific knowledge and the development of scalable benchmarking platforms to further refine the efficacy and reliability of LLM-EA systems.

\bibliographystyle{elsarticle-num}

\end{document}